\documentclass{article}

\PassOptionsToPackage{numbers, compress}{natbib}

\usepackage[final]{neurips_2024}

\usepackage[utf8]{inputenc} 
\usepackage[T1]{fontenc}    
\usepackage{hyperref}       
\usepackage{url}            
\usepackage{booktabs}       
\usepackage{amsfonts}       
\usepackage{nicefrac}       
\usepackage{microtype}      
\usepackage{xcolor}         
\usepackage{colortbl}
\usepackage{enumitem}

\usepackage{amssymb}
\usepackage{amsmath}
\usepackage{stmaryrd}
\usepackage{gensymb}

\usepackage{wrapfig}
\usepackage{graphicx}

\usepackage{multirow}

\usepackage[linesnumberedhidden,ruled,vlined]{algorithm2e}

\usepackage{caption}
\usepackage{subcaption}
\usepackage{todonotes}

\newcommand{\mymethod}{INTEGER}

\newcommand{\fgcmfull}{Feature-Geometry Coherence Mining}
\newcommand{\fgcmfullb}{\textbf{F}eature-\textbf{G}eometry \textbf{C}oherence \textbf{M}ining}
\newcommand{\fgcm}{FGCM}

\newcommand{\pbsafull}{Per-Batch Self-Adaption}
\newcommand{\fgcfull}{Feature-Geometry Clustering}

\newcommand{\mdsfull}{Mixed-Density Student}
\newcommand{\mdsfullb}{\textbf{M}ixed-\textbf{D}ensity \textbf{S}tudent}
\newcommand{\mds}{MDS}

\newcommand{\atlfull}{Anchor-Based Contrastive Learning}
\newcommand{\atlfullb}{\textbf{A}nchor-\textbf{B}ased \textbf{Cont}rastive Learning}
\newcommand{\atl}{ABCont}

\newcommand{\kitti}{KITTI}
\newcommand{\nuscnenes}{nuScenes}

\newcommand{\rgbd}{RGB\nobreakdash-D}

\newcommand{\scpcr}{SC\textsuperscript{2}-PCR}

\definecolor{mylightblue}{HTML}{D0E1ED}

\newcommand{\xkznew}[1]{\textcolor{black}{{#1}}}

\title{Mining and Transferring Feature-Geometry Coherence for Unsupervised Point Cloud Registration}

\author{
	Kezheng Xiong$^{ab}$, Haoen Xiang$^{ab}$, Qingshan Xu$^{c}$, Chenglu Wen$^{ab}$\thanks{Corresponding author} , Siqi Shen$^{ab}$, \\  \textbf{Jonathan Li}$^{d}$, \textbf{Cheng Wang$^{ab}$}\\
	$^{a}$Fujian Key Laboratory of Sensing and Computing for Smart Cities, \\ Xiamen University, China. \\
  $^{b}$Key Laboratory of Multimedia Trusted Perception and Efficient Computing, \\ Ministry of Education of China, Xiamen University, China. \\
  $^{c}$Nanyang Technological University, Singapore.\\
  $^{d}$University of Waterloo, Waterloo, Canada \\
	\texttt{\{xiongkezheng,haoenxiang\}@stu.xmu.edu.cn}\\
  \texttt{\{clwen,siqishen,cwang\}@xmu.edu.cn}\\
  \texttt{qingshan.xu@ntu.edu.sg}, \texttt{junli@uwaterloo.ca}\\
}

\begin{document}

\maketitle

\begin{abstract}
  Point cloud registration, a fundamental task in 3D vision, has achieved remarkable success with learning-based methods in outdoor environments.
  Unsupervised outdoor point cloud registration methods have recently emerged to circumvent the need for costly pose annotations.
  However, they fail to establish reliable optimization objectives for unsupervised training, 
  either relying on overly strong geometric assumptions,
  or suffering from poor-quality pseudo-labels due to inadequate integration of low-level geometric and high-level contextual information.
  We have observed that in the feature space, latent new inlier correspondences tend to cluster around respective positive anchors that summarize features of existing inliers.
  Motivated by this observation, we propose a novel unsupervised registration method termed \mymethod{} to incorporate high-level contextual information for reliable pseudo-label mining.
  Specifically, we propose the \fgcmfull{} module to dynamically adapt the teacher for each mini-batch of data during training and discover reliable pseudo-labels by considering both high-level feature representations and low-level geometric cues.
  Furthermore, we propose \atlfull{} to facilitate contrastive learning with anchors for a robust feature space.
  Lastly, we introduce a \mdsfull{} to learn density-invariant features, addressing challenges related to density variation and low overlap in the outdoor scenario.
  Extensive experiments on \kitti{} and \nuscnenes{} datasets demonstrate that our \mymethod{} achieves competitive performance in terms of accuracy and generalizability.
  \texttt{\textbf{\href{https://github.com/kezheng1204/INTEGER.git}{[Code Release]}}}
\end{abstract}

\section{Introduction}
\label{sec:intro}

Point cloud registration is a fundamental task in autonomous driving and robotics.
It aims to align two partially overlapping point clouds with a rigid transformation.
Learning-based methods have achieved remarkable success in outdoor point cloud registration\citep{cao2021pcam,lu2021hregnet,xiong2024speal,qin2022geometric,liu2023regformer}. 
PCAM\citep{cao2021pcam} pioneered the integration of low-level geometric and high-level contextual information, inspiring subsequent works\citep{lu2021hregnet,xiong2024speal,qin2022geometric,liu2023regformer}.
However, these supervised methods suffer from poor generalizability and reliance on costly pose annotations\citep{geiger2012we,caesar2020nuscenes,sun2020scalability}, 
underscoring the need for unsupervised methods to address these challenges in real-world applications.

Despite recent progress\citep{el2021unsupervisedr,yuan2023pointmbf,el2021bootstrap,liu2024extend} in unsupervised registration methods, the task remains challenging and underexplored, especially in outdoor scenarios where LiDAR point clouds are large-scale and complexly distributed.
Some methods\citep{el2021unsupervisedr,yuan2023pointmbf,el2021bootstrap} optimize photometric and depth consistency, 
limiting their applicability to indoor scenarios where \rgbd{} data and differentiable rendering are feasible. 
Others\citep{shen2022reliable,shi2023overlapbiasmatchingnecessary} learn global alignment and neighborhood consensus, 
but struggle with low overlap and density variation in outdoor settings.
Recent advances resort to pseudo-label-based frameworks, achieving promising results in outdoor scenarios\citep{liu2024extend,yang2021sgp}.
However, they rely solely on geometric cues to mine and filter pseudo-labels, neglecting the complementarity of high-level contextual information in feature space.
Their partiality results in incomplete scene perception, leading to noisy and suboptimal optimization objectives.

\begin{figure}[tb]
  \centering
  \vspace{-0.5cm}
  \includegraphics[width=\textwidth]{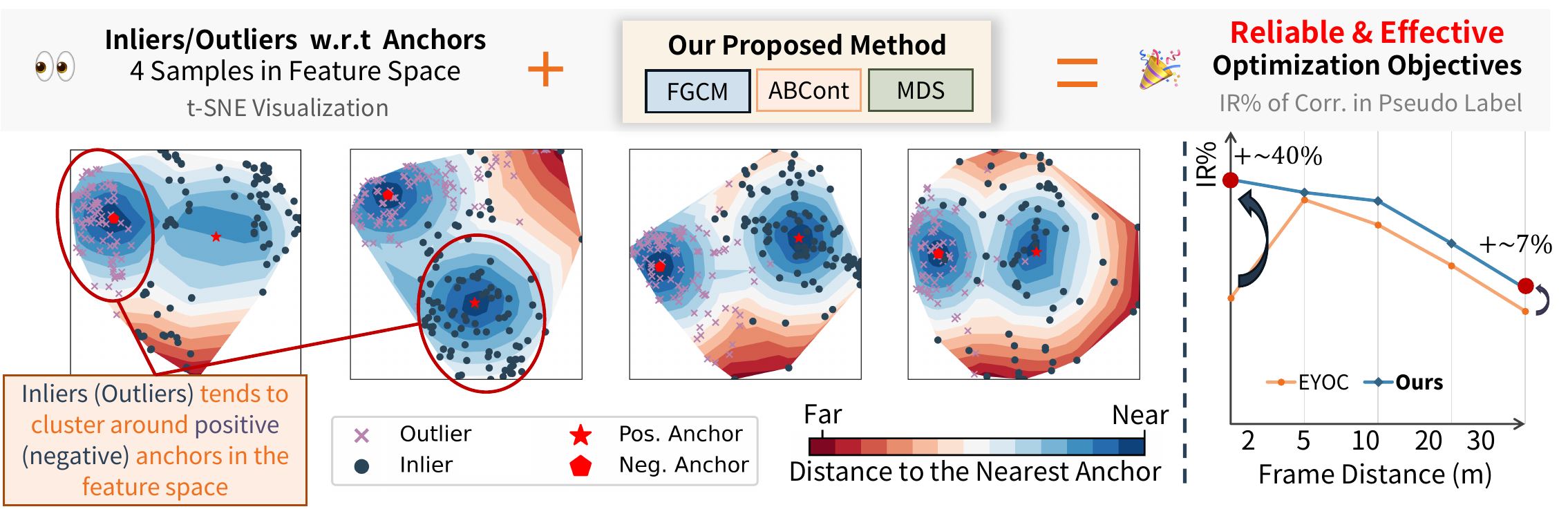}
  \vspace{-0.5cm}
  \caption{
    (1) \textbf{Motivation}: new inliers (outliers) tend to cluster around latent \emph{positive (negative) anchors} that represent existing inliers (outliers) in the feature space, respectively. 
    (2) \textbf{Performance}: pseudo-labels from \mymethod{} are more robust and accurate than the previous state-of-the-art EYOC\citep{liu2024extend}.
  }
  \label{fig:teaser} 
  \vspace{-0.5cm}
\end{figure}

Various 2D \citep{chen2018encoder,lin2017feature} and 3D vision tasks \citep{lang2019pointpillars,shi2020pv,hu2021learning,lu2021hregnet,liu2023regformer,qin2022geometric} have benefited from integrating \emph{both} low-level \emph{and} high-level information.
In point cloud registration, as illustrated in Fig.~\ref{fig:teaser} (\textbf{Left}), we observe that potential inliers (outliers) tend to cluster around \emph{positive (negative) anchors} that summarize the features of existing inliers (outliers) in the feature space, respectively.
This suggests that high-level contextual information is adept at discovering inliers from a global perspective of the scene. 
Meanwhile, low-level geometric cues have proven effective in rejecting outliers\citep{shen2022reliable,chen2022sc2,zhang2024fastmac,zhang20233d}. 
Inspired by this, we propose a novel method, termed \textbf{\mymethod},
which adopts a \emph{teacher-student framework} to m\textbf{IN}e and \textbf{T}ransfer f\textbf{E}ature-\textbf{GE}ometry cohe\textbf{R}ence for unsupervised point cloud registration.

Specifically, our method starts by initializing a teacher with synthetic pairs generated from each
point cloud scan, and then transfers to real point cloud pairs with a teacher-student framework. Building upon our observations, we introduce the \fgcmfullb{} (\fgcm) module for the \emph{teacher}, which first adapts the teacher to each mini-batch of real data to establish a denoised feature space.
Reliable pseudo-labels, including correspondences and anchors, are then generated based on our key observation by iteratively mining potential inliers based on their similarity to anchors and rejecting outliers via spatial compatibility \citep{chen2022sc2}. 
These robust pseudo-labels mined by \fgcm{} not only accurately include inlier correspondences as shown in Fig.~\ref{fig:teaser} (\textbf{Right}), but also aggregate effective representations of inliers and outliers from the teacher. 
We refer to this characteristic as \emph{feature-geometry coherence}.
To further enhance robustness and transfer feature-geometry coherence to the student, we propose \atlfullb{} (\atl{}) for contrastive learning with anchors. 
Meanwhile, we design a succinct and efficient \mdsfullb{} (\mds) for the \emph{student} to learn density-invariant features using teacher's anchors, 
overcoming density variation and low overlap in distant scenarios.

We extensively evaluate our method on two large-scale outdoor datasets, \kitti{} and \nuscnenes{}.
By exploiting feature-geometry coherence for reliable optimization objectives, \mymethod{} outperforms existing unsupervised methods by a considerable margin.
It even performs competitively compared to state-of-the-art supervised methods, especially in distant scenarios. 
To the best of our knowledge, our approach is the first to integrate both low-level and high-level information for producing pseudo-labels of unsupervised point cloud registration. 
Overall, our contributions are threefold:
\begin{itemize}[leftmargin=10pt]
  \item  We propose \mymethod, a novel method to exploit low-level and high-level information for unsupervised point cloud registration, achieving superior performance in complex outdoor scenarios.
  \item We introduce \fgcm{} and \mds{} for the teacher and student, respectively, to mine reliable pseudo-labels and learn density-invariant features.
  \item We design \atl{} to mitigate pseudo-label noise and facilitate contrastive learning with anchors for a robust feature space.
\end{itemize}

\section{Related Works}

\paragraph{Supervised Registration.}
There are two categories of supervised registration approaches:
Correspondence-based methods \cite{choy2019fully,deng2018ppf,gojcic2019perfect,qin2022geometric,yew2022regtr,yu2023peal,xiong2024speal} first extract point correspondences and then estimate the transformation with robust pose estimators.
In contrast, direct registration methods \cite{xu2021omnet,aoki2019pointnetlk,huang2020feature} extract global feature vectors and regress the transformation directly with a neural network. 
Recently, a series of works \cite{liu2024extend,liu2023apr} have tackled distant point cloud registration, which is crucial for real-world applications.

\paragraph{Unsupervised Registration.}
Previous researches in unsupervised registration mainly focus on indoor scenes.
BYOC\citep{el2021bootstrap} suggests that random 2D CNNs generate robust image correspondences for supervising 3D registration networks.
Meanwhile, render-based methods\citep{el2021unsupervisedr,yuan2023pointmbf} leverage differentiable renders as the supervision signal of 3D registration.
However, these methods are restricted to \rgbd{} input.
To address this, \citet{mei2023unsupervised} enforce consistencies between 
\xkznew{Gaussian Mixture Models}
for unsupervised training, using only point cloud as input. 
\citet{shen2022reliable} introduce an inlier evaluation method based on neighborhood consensus.
However, its performance drops when the overlap is low.
SGP\citep{yang2021sgp} proposes a teacher-student framework for self-supervised learning from hand-crafted feature descriptors.
EYOC\citep{liu2024extend} introduces progressive training and spatial filtering to adapt the model to distant point cloud pairs gradually, demonstrating promising results in outdoor scenarios.

\paragraph{Robust Pose Estimators.}
Pose estimators evaluate inliers and estimate poses from input correspondence sets.
Traditional methods such as RANSAC\citep{fischler1981random} suffer from inefficiencies.
Learning-based methods\citep{bai2021pointdsc,choy2020deep,lee2021deep} learn to predict inliers and poses using neural networks.
However, they require training and are thus constrained to supervised settings.
To address this, non-parametric methods have emerged. \citet{chen2022sc2} introduced \(SC^2\)-measurements for robust inlier selection.
Graph-based methods such as MAC\citep{zhang20233d} and FastMAC\citep{zhang2024fastmac} approximate maximal cliques for fast and accurate inlier evaluation.

\section{Methodology}
\label{sec:method}

\paragraph{Problem Formulation.}
Given two point clouds \(\mathcal{P}\!=\!\{\mathbf{p}_i\}\in\mathbb{R}^{m\times 3}\) and \(\mathcal{Q}\!=\!\{\mathbf{q}_j\}\in\mathbb{R}^{n\times 3}\), the goal of point cloud registration is to uncover the rigid transformation \(\mathbf{T}=\{\mathbf{R},\mathbf{t}\}\) that perfectly aligns \(\mathcal{P}\) to \(\mathcal{Q}\), where \(\mathbf{R}\in\mathrm{SO}(3)\) is the rotation matrix and \(\mathbf{t}\in\mathbb{R}^{3}\) is the translation vector.
When the two point clouds are acquired at a large distance $d$ such as when \(d\in[5\mathrm{m}, 50\mathrm{m}]\),
the registration task faces the challenges of low overlap and density variation \citep{liu2023apr,liu2024extend,liu2023density}. 
Therefore, it is crucial to learn density-invariant features.

\begin{figure}
  \vspace{-0.3cm}
  \centering 
  \includegraphics[width=1.0\textwidth]{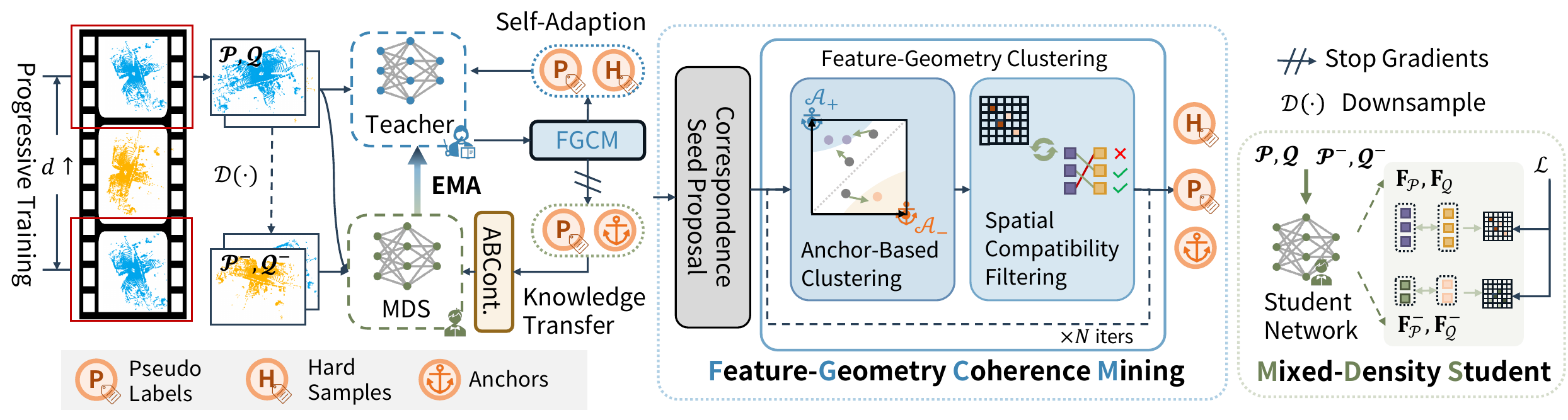} 
  \vspace{-0.5cm}
  \caption{
  \textbf{The Overall Pipeline.} \fgcm(Sec.~\ref{subsec:fgcm}) first adapt the teacher model to a \emph{data-specific teacher} for the current mini-batch, and then mine reliable pseudo-labels.
  Next, \mds(Sec.~\ref{subsec:mds}) learns density-invariant features from pseudo-labels.
  \atl(Sec.~\ref{subsec:abcont}) is applied for adapting the teacher and transferring knowledge to the student in the feature space.
  }
  \label{fig:pipeline} 
\end{figure}
\paragraph{Overall Pipeline.}
\mymethod{} adopts a two-stage training scheme and a teacher-student framework.
Training of \mymethod{} consists of two stages: First, we initialize the teacher with synthetic data.
Then, we train a student model on \emph{real data} with the reliable pseudo-labels mined by the teacher.
The overall pipeline and proposed modules are illustrated in Fig.~\ref{fig:pipeline}. During teacher-student training, \fgcm{} first dynamically adapts the teacher model \(\theta\) to a \emph{data-specific teacher} \(\phi\) designated for the current mini-batch, and then mines reliable pseudo-labels with the adapted teacher.
Next, the \mds{} learns density-invariant features by learning to match regular and sparse views of point cloud pairs supervised by pseudo-labels mined by \(\phi\).
A pseudo-label \(\mathcal{I} = \{\mathcal{C},\hat{\mathcal{C}},\mathcal{A}_+,\mathcal{A}_-\}\) contains correspondences  \(\mathcal{C},\hat{\mathcal{C}}\) to supervise dense matches and sparse matches, respectively. 
The feature-space positive and negative anchors, denoted respectively by \(\mathcal{A}_+\) and \(\mathcal{A}_-\), serve as overall representatives of inliers and outliers in the feature space.
For a correspondence \(^{(i,j)}\mathcal{C}=(\mathbf{p}_i,\mathbf{q}_j)\in\mathcal{C}\), the correspondence features are defined as \(\mathbf{F}_\mathcal{C}^{(i,j)} = \mathbf{F}^\mathcal{P}_i-\mathbf{F}^\mathcal{Q}_j\).
Then, the positive and negative anchors \(\mathcal{A}_+,\mathcal{A}_-\) are computed as the average of the respective features of inliers \(\mathcal{C}_+\) and outliers \(\mathcal{C}_-\)
\begin{equation}
  \label{eqn:anchorcomputation}
  \begin{aligned}
    \mathcal{A}_+ &= \frac{1}{|\mathcal{C}_+|}\sum_{(\mathbf{p}_i,\mathbf{q}_j)\in\mathcal{C}_+}\!\!\!\!\!\!\mathbf{F}_\mathcal{C}^{(i,j)}, \
    \mathcal{A}_- &= \frac{1}{|\mathcal{C}_+|}\sum_{(\mathbf{p}_i,\mathbf{q}_j)\in\mathcal{C}_-}\!\!\!\!\!\!\mathbf{F}_\mathcal{C}^{(i,j)}, 
  \end{aligned}
\end{equation}
\atl{} is applied to effectively learn a robust student guided by anchors from the teacher.
Progressive training \citep{liu2024extend} is adopted to gradually train the student to adapt to pairs of distant point clouds.

\subsection{Synthetic Teacher Initialization}
To initialize a teacher model, \citet{liu2024extend} assume that two consecutive frames approximately have no relative transformation and pretrain the teacher with the \emph{identity} transformation.
However, the errors introduced in such approximation lead to suboptimal initial teachers.
To address this, inspired by existing efforts\citep{horache20213d,xie2020pointcontrast}, we instead pretrain the teacher with synthetic pairs generated from each real scan.
Specifically, we follow PointContrast\citep{xie2020pointcontrast} to generate two partially overlap fragments for each scan.
We additionally apply periodic sampling\citep{horache20213d} to remove points periodically with respect to a random center, simulating the irregular sampling of LiDAR.
Please refer to the Appendix for a visualization of synthetic pairs.

\subsection{\fgcmfull}
\label{subsec:fgcm}

\begin{wrapfigure}{r}{0.6\textwidth}
  \centering 
  \vspace{-0.5cm}
  \includegraphics[width=0.6\textwidth]{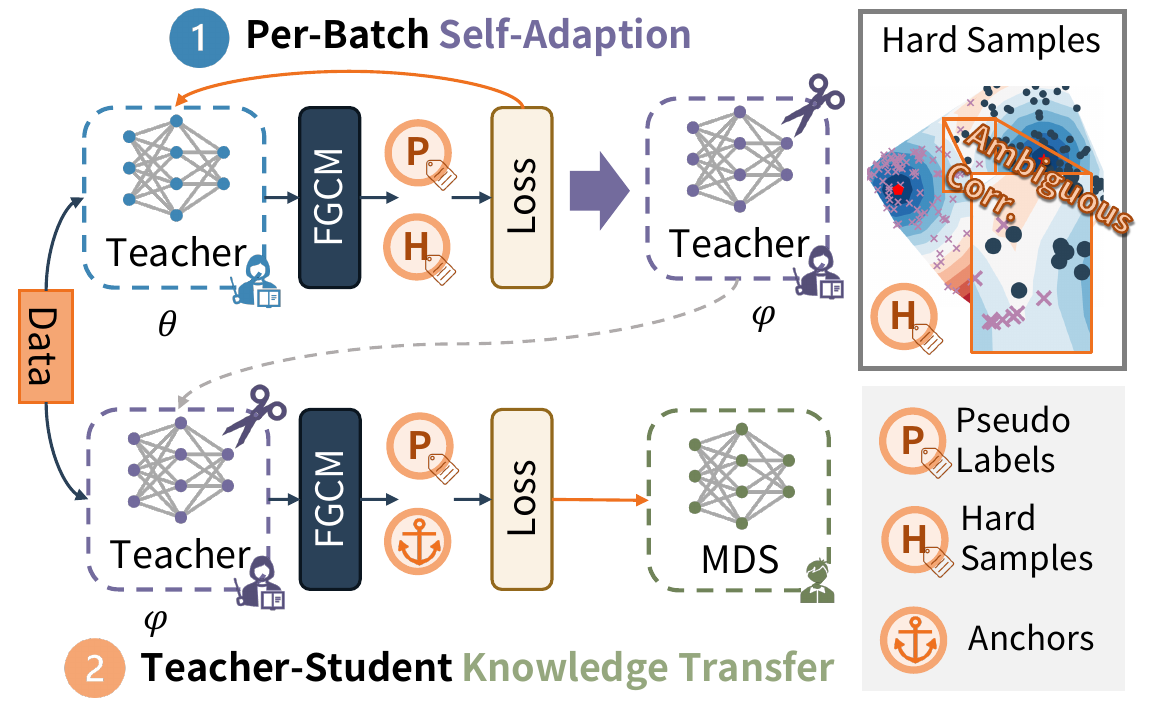} 
  \vspace{-0.4cm}
  \caption{The two-pass usage of the proposed \fgcm.}  
  \vspace{-0.2cm}
  \label{fig:fcem} 
\end{wrapfigure}

With the teacher initialized on synthetic pairs, our goal is to provide reliable supervision for the student.
Despite efforts to ensure an effective initialization, a distribution discrepancy persists between synthetic and real data.
Hence, we introduce a \emph{train-only} \fgcm.
As is depicted in Fig.~\ref{fig:pipeline}, \fgcm{} starts with Correspondence Seed Proposals for \(\mathcal{C}^0\) using a simple similarity threshold.
Subsequently, Feature-Geometry Clustering extends from \(\mathcal{C}^0\) by mining additional reliable correspondences and anchors, which serve as effective optimization objectives.

As is illustrated in Fig.~\ref{fig:fcem}, for each mini-batch, we use \fgcm{} in a two-pass manner. 
In the first forward pass, we perform Per-Batch Self-Adaption on the teacher model \(\theta\) to establish a denoised feature space, yielding a data-specific teacher \(\phi\).
In the second forward pass, the adapted teacher \(\phi\) and \fgcm{} are used to mine reliable pseudo-labels \(\mathcal{I}\), which are then used to train the student, achieving Teacher-Student Knowledge Transfer.

\paragraph{Feature-Geometry Clustering.}
\emph{Feature-Geometry Clustering} is central to the \fgcm{}, designed to extend initial correspondence proposals by integrating both high-level \emph{feature} representations and low-level \emph{geometric} cues.
It iteratively includes speculative inliers based on feature-space clustering, followed by outlier rejection with spatial compatibility filtering.
We empirically adopt \scpcr \citep{chen2022sc2} for spatial compatibility filtering. 
Our experiments show that our method is agnostic to the choice of spatial compatibility measures.

To discover latent correspondences in the feature space, it is necessary to measure the similarity between putative correspondences and anchors.
Inspired by \citet{xia2023coin}, we compute the feature similarity using both Euclidean distance and cosine similarity.
Specifically, for a correspondence \(\left(\mathbf{p}_i,\mathbf{q}_j\right)\in \mathcal{C}\) with its features \(\mathbf{F}_\mathcal{C}^{(i,j)}\),
The similarity \(S_c^+\) and \(S_c^-\) w.r.t. respective anchors \(\mathcal{A}_+\) and \(\mathcal{A}_-\) is computed as:
\begin{equation}
  \label{eqn:corrsimilarity}
  S^+_c = \mathrm{min}\{\mathrm{D_\text{E}}(\mathcal{A}_+, \mathbf{F}_\mathcal{C}^{(i,j)}),\mathrm{D}_\text{C}(\mathcal{A}_+, \mathbf{F}_\mathcal{C}^{(i,j)})\}, \ 
  S^-_c = \mathrm{min}\{\mathrm{D_\text{E}}(\mathcal{A}_-, \mathbf{F}_\mathcal{C}^{(i,j)}),\mathrm{D}_\text{C}(\mathcal{A}_-, \mathbf{F}_\mathcal{C}^{(i,j)})\},
\end{equation}
where \(\mathrm{D}_\text{E}(\mathbf{F}_1,\mathbf{F}_2)\!\!=\!\!1\!\!-\!\!\mathrm{min}(L_2(\mathbf{F}_1,\mathbf{F}_2),1)\) and \(\mathrm{D}_\text{C}(\mathbf{F}_1,\mathbf{F}_2)\!=\!\left(\mathrm{cos}(\mathbf{F}_1,\mathbf{F}_2)\!+\!1\right)/2\) are \emph{normalized} Euclidean distance and cosine similarity, respectively.

\begin{algorithm}[tb]
  \caption{Feature-Geometry Clustering}
  \label{alg:fgce}
  \DontPrintSemicolon
  \KwIn{Initial correspondence seed proposals $\mathcal{C}^0$}
  Compute initial $\mathbf{U}^\mathcal{P}$, $\mathbf{U}^\mathcal{Q}$ and anchors $\mathcal{A}_+$, $\mathcal{A}_-$ with Eq.~\ref{eqn:anchorcomputation}\;
  \For{$i=1$ to $max\_iters$}
  {
    Generate unclassified correspondences  $\mathcal{C}_\text{U} \gets \mathrm{FeatureMatching}\left(\mathbf{U}^\mathcal{P}, \mathbf{U}^\mathcal{Q}\right)$ \;
    Select $\mathcal{C}_\text{U}^{\text{top-}k}$ with top-$k$ $S^+_c$ satisfying $S^+_c>S^-_c$ based on Eq.~\ref{eqn:corrsimilarity}\;
    Update $\mathcal{C}^{i} \gets \mathcal{C}^{i-1} \cup \mathcal{C}_\text{U}^{\text{top-}k}$ \tcp*{Anchor-Based Clustering}
    Filter $\mathcal{C}^i$ with spatial compatibility to produce $\mathcal{C}^i_+$, $\mathcal{C}^i_-$ \tcp*{Spatial Compatibility Filtering}
    Update $\mathbf{U}^\mathcal{P}$, $\mathbf{U}^\mathcal{Q}$ and $\mathcal{A}_+$, $\mathcal{A}_-$ with Eq.~\ref{eqn:anchorcomputation}\;
    \If{$\left|\mathcal{C}^i_+\right|=\left|\mathcal{C}^{i-1}_+\right|$ or $\left|\mathcal{C}^i_-\right|=\left|\mathcal{C}^{i-1}_-\right|$}
    {
      $\mathcal{C}\gets \mathcal{C}^i_+$\;
      \textbf{break}\;
    }
  }
  \KwRet{$\mathcal{C}$, $\mathcal{A}_+$, $\mathcal{A}_-$}\;
\end{algorithm}

The algorithm is detailed in Alg.~\ref{alg:fgce}.
Given correspondence set \(\mathcal{C}^i\) at \(i\)-th iteration,  
we define unclassified points \(\mathbf{U}^\mathcal{P}, \mathbf{U}^\mathcal{Q}\) in \(\mathcal{P}\) and \(\mathcal{Q}\) as:
\begin{equation}
  \label{eqn:pointspartition}
  \mathbf{U}^\mathcal{P}=\left\{\mathbf{p}|\mathbf{p}\in \mathcal{P} \wedge (\mathbf{p},*) \notin \mathcal{C}^i   \right\}, \ 
  \mathbf{U}^\mathcal{Q}=\left\{\mathbf{q}|\mathbf{q}\in \mathcal{Q} \wedge (*,\mathbf{q}) \notin \mathcal{C}^i   \right\}
\end{equation} 
Then, the algorithm takes an iterative approach, starting from the given initial correspondence set \(\mathcal{C}^0\): 
during the \(i^\text{th}\) iteration, it (1) generates putative correspondences from \(\mathbf{U}^\mathcal{P}\) and \(\mathbf{U}^\mathcal{Q}\) via feature matching;
(2) expands \(\mathcal{C}^{i-1}\) with top-\(k\) similar correspondences to positive anchors measured by \(\mathcal{S}_c\), yielding \(\mathcal{C}^i\);
(3) filters the expanded correspondence set with spatial compatibility and updates the anchors based on Eq.~\ref{eqn:anchorcomputation};
(4) updates \(\mathbf{U}^\mathcal{P}, \mathbf{U}^\mathcal{Q}\) and \(\mathcal{A}_+$, $\mathcal{A}_-\) according to Eq.~\ref{eqn:pointspartition}.
The iteration stops when the number of inliers and outliers converges, or the maximum iteration is reached.

With ample accurate correspondences included in \(\mathcal{C}\), we can then estimate a more accurate transformation \(\mathbf{T}\) and compute \(\hat{\mathcal{C}}\) using nearest neighbor search (NN-search) with \(\mathbf{T}\).
We do not directly apply Alg.~\ref{alg:fgce} for sparse pairs because, in downsampled views, the features become less descriptive \citep{liu2023density}, hindering feature-based approaches.

\paragraph{Per-Batch Self-Adaption.}
Throughout the iterations in Alg.~\ref{alg:fgce}, positive and negative anchors gradually aggregate representative and discriminative features of inliers and outliers, respectively.
In the first forward pass, noise exists due to distributional discrepancies, leading to the rejection of some correspondences by spatial compatibility. 
These rejected correspondences are \emph{hard samples}: ambiguous correspondences that are closer to positive anchors in the feature space but are more likely to be outliers. 
We leverage these hard samples for teacher self-adaptation by applying the InfoNCE\citep{oord2018representation} loss, guiding the teacher to distinguish them from the positive anchors. 
This step results in the adapted teacher \(\phi\), which produces more discriminative features for the current mini-batch.

Focusing only on hard samples for self-adaptation is more efficient than simply using all correspondences for self-adaptation.
Hard samples capture the key ambiguities in the feature space while introducing only a limited number of pairwise relationships. 
This defines a clear and reliable optimization objective for self-adaptation. 
In contrast, self-adaption with all correspondences not only slows down the training,
but also introduces too many already-distinguishable pairwise relationships, 
diluting the focus on feature-space ambiguity and thus hindering effective adaptation.

\paragraph{Teacher-Student Knowledge Transfer.}
After Per-Batch Self-Adaption, the feature space of the adapted teacher \(\phi\) is expected to contain less noise.
Consequently, the positive and negative anchors become sufficiently representative and discriminative now, enabling effective guidance for the student to learn a robust feature space.
For Teacher-Student Knowledge Transfer, we utilize both the correspondences and anchors from the adapted teacher \(\phi\) to train the student using the proposed \atl. 
Unlike existing methods\citep{yang2021sgp,liu2024extend} that rely solely on correspondences, our approach directly bridges the teacher and student in the feature space via anchors, providing a clear and effective optimization objective for the student.

\subsection{\atlfull}
\label{subsec:abcont}

\begin{wrapfigure}{r}{0.4\textwidth}
  \centering
  \vspace{-0.8cm}
  \includegraphics[width=0.4\textwidth]{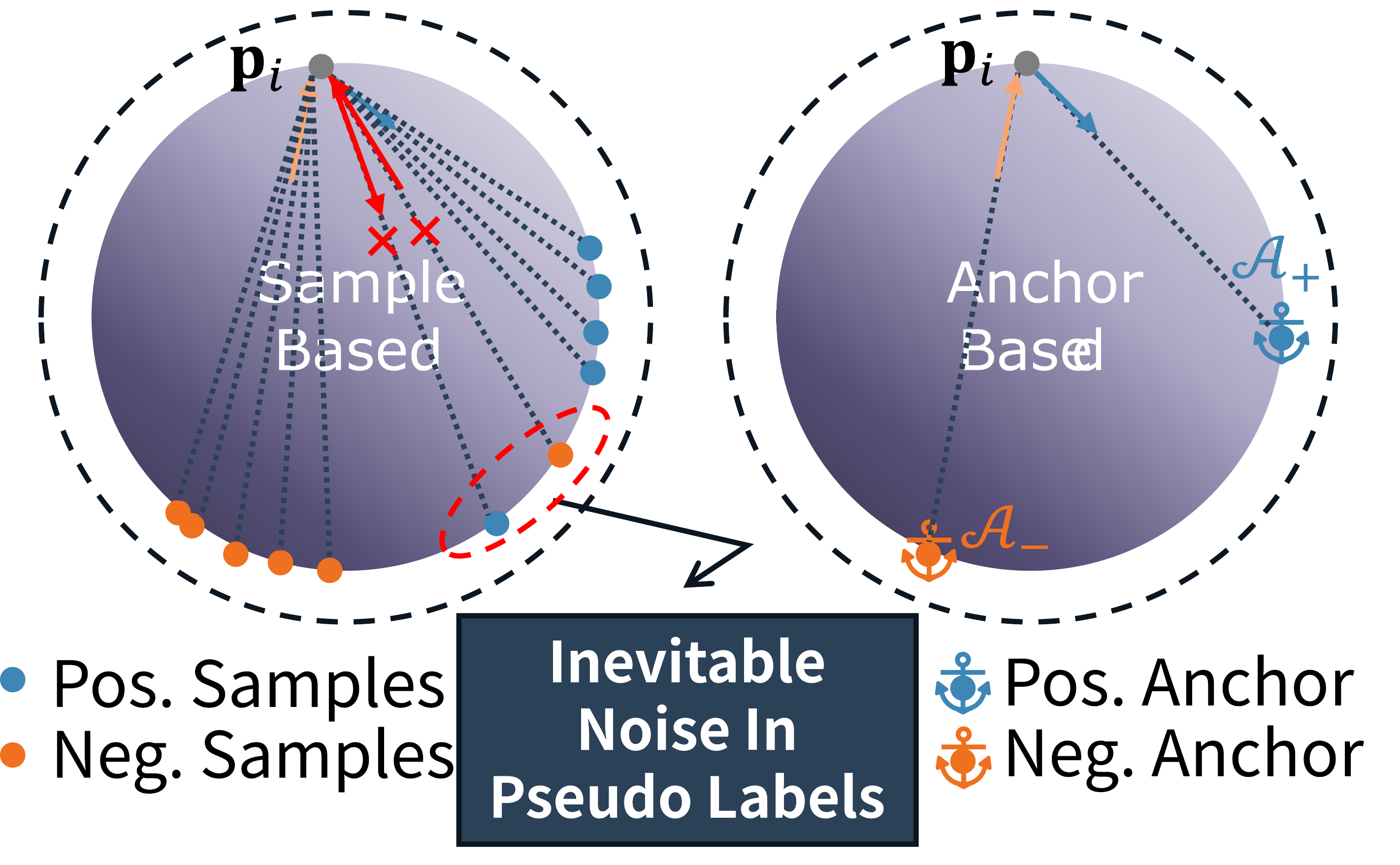}
  \vspace{-0.5cm}
  \caption{\textbf{Toy Example for \atl.} Anchor-based methods introduce fewer pairwise relationships and are robust against inevitable label noise.}
  \label{fig:anchormotivation}
  \vspace{-0.5cm}
\end{wrapfigure}
Contrastive learning has been widely adopted to train registration models\citep{huang2021predator,choy2019fully,qin2022geometric,xiong2024speal,hu2021learning}.
Recently, a surge of research on various tasks involves \emph{anchor-based} or \emph{proxy-based} approaches to facilitate contrastive learning due to their robustness against inconsistency and noise in the feature space\citep{bai2022dual,barbany2024procsim,xia2023coin}, superiority in generalizability\citep{yao2022pcl} and ability to learn discriminative features\citep{chen2022geometry}.
Therefore, we design \atl{} to leverage positive and negative anchors to facilitate effective contrastive learning with the pseudo-labels, where noise and outliers are inevitable.
As shown in Fig.~\ref{fig:anchormotivation}, with anchor-based representations, \atl{} sets up a convergence target that is more robust against label noise.
Moreover, it is more efficient because the number of additionally-introduced pairwise relationships is reduced\citep{yao2022pcl}.

Specifically, we propose the \atl{} loss \(\mathcal{L}_\text{\atl}\! = \!\mathcal{L}_\text{reg}\! +\! \lambda_\text{corr} \mathcal{L}_\text{corr}\), where \(\mathcal{L}_\text{corr}\) is the anchor-based correspondence loss,
weighted by a hyperparameter \(\lambda_\text{corr}\) to complement the registration loss \(\mathcal{L}_\text{reg}\) originally used by the feature extractors.
The student's feature matching results can be classified into inliers \(\mathcal{C}_+\) and outliers \(\mathcal{C}_-\) based on the pseudo-labels from the teacher.
Then, anchors \(\{\mathcal{A}_+,\mathcal{A}_-\}\) from the teacher are designated as a \emph{universal inlier} and a \emph{universal outlier}, resulting in augmented inliers and outliers:
\begin{equation}
  \label{eqn:anchorcorrespondence}
  \begin{aligned}
    \mathcal{C}^\star_+ &= \mathcal{C}_+ \cup \mathrm{sg}(\mathcal{A}_+), \ \mathcal{C}^\star_- &= \mathcal{C}_- \cup \mathrm{sg}(\mathcal{A}_-),
  \end{aligned}
\end{equation}
where \(\mathrm{sg}(\cdot)\) denotes the stop-gradient operator, preventing gradients from flowing back to the teacher.
Following existing efforts\citep{xia2023coin,xie2020pointcontrast}, we sample \(n_p\) correspondences randomly and formulate \(\mathcal{L}_\text{aux}\) as a contrastive learning problem to distinguish inliers from outliers.
InfoNCE\citep{oord2018representation} loss is then applied to these correspondence features:
\begin{equation}
  \label{eqn:auxloss}
  \begin{aligned}
    \mathcal{L}_\text{corr} = -\frac{1}{n_p}\sum_{i=1}^{n_p}\log\frac{\exp(\beta_p^i)}{\exp(\beta_p^i)+\sum_{j=1}^{n_n}\exp(\beta_n^j)},
  \end{aligned}
\end{equation}
where \(\beta_p^i\) and \(\beta_n^j\) are the distance between the \(i^\text{th}\) positive correspondence and the \(j^\text{th}\) negative correspondence, respectively.
\atl{} is pivotal in transferring feature-geometry coherence from the teacher to the student: the accurate pseudo-labels for correspondences, combined with anchors, 
enable the student to learn discriminative features efficiently.
Anchors from the teacher impose direct constraints on the student's feature space, encouraging the student to replicate the teacher's feature-space matchability. 
This leads to a more effective transfer of feature-geometry coherence. 

\subsection{\mdsfull}
\label{subsec:mds}

The density of LiDAR point clouds varies greatly with the distance to the sensor, posing challenges for matching distance point clouds effectively\citep{liu2023density}.
To address this, it is crucial for a student model to learn density-invariant features, ensuring robust correspondences across varying point densities.
Previous methods \citep{liu2023apr,liu2023density} have sought density invariance through auxiliary reconstruction tasks or by identifying positive groups, 
but these techniques are either computationally expensive or depend on precise supervision.
\citet{xia2024hinted} introduced a simple yet effective technique for density invariance in object detection by using features from downsampled views.
Inspired by this, we propose \mdsfull{} to learn density-invariant features from reliable correspondences.

Specifically, given point clouds \(\mathcal{P}\) and \(\mathcal{Q}\), we compute their sparsely-downsampled views \(\mathcal{P}^{-}\) and \(\mathcal{Q}^{-}\) with increased voxel sizes.
We then extract student's features \((\mathbf{F}_\mathcal{P},\mathbf{F}_\mathcal{Q})\) and \((\mathbf{F}^{-}_\mathcal{P},\mathbf{F}^{-}_\mathcal{Q})\).
Using features from point clouds of different density, we compute \emph{dense matches} and \emph{sparse matches} through matching respective features.
We then apply \atl{} to both sets of matches, 
encouraging the extraction of similar features at corresponding spatial locations across point clouds of varying densities, thereby promoting density-invariant feature learning.

\paragraph{Loss Aggregation.} 
The student's overall training loss is aggregated as a weighted combination of \(\mathcal{L}_{\text{\atl}}\) on both dense and sparse matches:
\begin{equation}
  \mathcal{L} = \mathcal{L}^{(\mathcal{P},\mathcal{Q})}_\text{\atl} + \lambda_1 \mathcal{L}^{(\mathcal{P}^{-},\mathcal{Q}^{-})}_\text{\atl} , 
\end{equation}
where \(\lambda_1\) is a weight for the sparse match.

\section{Experiments}
\label{sec:expr}

We mainly evaluate \mymethod{} on two challenging public datasets: \kitti\cite{geiger2012we} and \nuscnenes\cite{caesar2020nuscenes}.
Both datasets adhere to official splits. 
The evaluation protocol follows the standard setting of EYOC\cite{liu2024extend}.
Please refer to the appendix for more details of implementation and experimental settings.
\paragraph{Metrics}
Following previous works\cite{qin2022geometric,lu2021hregnet,huang2021predator,liu2023density}, we evaluate the registration performance using \emph{Relative Rotation Error} (RRE), \emph{Relative Translation Error} (RTE) and \emph{Registration Recall} (RR).
Related to the practical purpose of outdoor registration, we additionally report \(\mathrm{RR}@\left[d_1,d_2\right)\) and \emph{mean Registration Recall}(mRR).
\(\mathrm{RR}@\left[d_1,d_2\right)\) is registration recall w.r.t pairs with distance \(d\in\left[d_1,d_2\right)\), following \cite{liu2024extend}.
mRR is defined as the average of \(\mathrm{RR}@\left[d_1,d_2\right)\) for all \(\left[d_1,d_2\right)\).
To measure the quality of correspondences in pseudo-labels, we report \emph{Inlier Ratio}(IR) of the \emph{\textbf{t}eacher} in the first epoch, denoted ``\textbf{t}IR@1\textsuperscript{st} Epoch''.

\paragraph{Baselines}
For supervised methods, We compare \mymethod{} with  FCGF\citep{choy2019fully}, Predator\citep{huang2021predator}, SpinNet\citep{ao2021spinnet}, D3Feat\citep{bai2020d3feat}, CoFiNet\citep{yu2021cofinet}, and Geometric Transformer(GeoTrans.)\citep{qin2022geometric}. 
For unsupervised methods, we compare with RIENet\citep{shen2022reliable} and EYOC\citep{liu2024extend}.
Following \citet{liu2024extend}, we report a variant of FCGF denoted as FCGF+C, which is FCGF trained with progressive training \citep{liu2024extend}.

\begin{table}[htb]
  \centering
  \renewcommand{\arraystretch}{0.9}
  \caption{
  \textbf{Comparisons with State-of-the-Art Methods.} ``\checkmark'' in the column ``U'' denotes the methods are \emph{\textbf{U}nsupervised}. 
  Otherwise, they are supervised. 
  The best \emph{unsupervised} results are highlighted in \textbf{bold}.
  ``\kitti \textrightarrow \nuscnenes''denotes generalizability results from \kitti{} to \nuscnenes.}
  \label{tbl:overallperf}
  \begin{tabular}{llccccccc}
    \toprule
    \multirow{2}{*}{Dataset}  & \multirow{2}{*}{Method} & \multirow{2}{*}{U} & \multirow{2}{*}{mRR} & \multicolumn{5}{c}{RR\(@ d\in\)}  \\
    \cmidrule{5-9}
    & & & & \(\left[5, 10\right)\) & \(\left[10, 20\right)\) & \(\left[20, 30\right)\) & \(\left[30, 40\right)\) & \(\left[40, 50\right)\) \\
    \midrule
    \multirow{8}{*}{\kitti} & FCGF & -- & 77.4 & 98.4 & 95.3 & 86.8 & 69.7 & 36.9 \\
    &FCGF+C & -- & 84.6 & 100.0 & 97.5 & 90.1 & 79.1 & 56.3 \\
    &Predator & -- & 87.9 & 100.0 & 98.6 & 97.1 & 80.6 & 63.1 \\
    &SpinNet & -- & 39.1 & 99.1 & 82.5 & 13.7 & 0.0 & 0.0 \\
    &D3Feat & -- & 66.4 & 99.8 & 98.2 & 90.7 & 38.6 & 4.5 \\
    &CoFiNet & -- & 82.1 & 99.9 & 99.1 & 94.1 & 78.6 & 38.7 \\
    &GeoTrans. & -- & 42.2 & 100.0 & 93.9 & 16.6 & 0.7 & 0.0 \\
    \cmidrule{2-9}
    & EYOC & \checkmark & 83.2 & \textbf{99.5} & 96.6& 89.1 & 78.6 & 52.3 \\
    & RIENet & \checkmark & 50.7 & 96.3 & 72.1 & 38.2 & 24.4 & 22.6 \\
    & \cellcolor{mylightblue} Ours & \cellcolor{mylightblue} \checkmark & \cellcolor{mylightblue} \textbf{84.0} & \cellcolor{mylightblue} \textbf{99.5} & \cellcolor{mylightblue} \textbf{97.1} & \cellcolor{mylightblue} \textbf{89.6} & \cellcolor{mylightblue} \textbf{79.6} & \cellcolor{mylightblue} \textbf{54.2}\\
    \midrule
    \multirow{5}{*}{\nuscnenes} & FCGF & -- & 39.5 & 87.9 & 63.9 & 23.6 & 11.8 & 10.2  \\
    &FCGF+C & -- & 59.3 & 96.2 & 85.1 & 59.6 & 35.8 & 20.0 \\
    &Predator & -- & 51.0 & 99.7 & 72.2 & 52.8 & 16.2 & 14.3 \\
    \cmidrule{2-9}
    & EYOC & \checkmark & 61.7 & 96.7 & 85.6 & 61.8 & 37.5 & 26.9 \\
    & RIENet & \checkmark & 47.1 & 96.5 & 57.9 & 36.6 & 25.8 & 18.9 \\
    & \cellcolor{mylightblue} Ours & \cellcolor{mylightblue} \checkmark &  \cellcolor{mylightblue} \textbf{63.1} & \cellcolor{mylightblue} \textbf{97.1} & \cellcolor{mylightblue} \textbf{86.9} & \cellcolor{mylightblue} \textbf{62.9} & \cellcolor{mylightblue} \textbf{39.6} & \cellcolor{mylightblue} \textbf{29.4} \\
    \midrule
    \multirow{3}{*}{\shortstack[c]{\kitti \\ \textdownarrow \\ \nuscnenes}} & EYOC & \checkmark & 55.3 & 96.2 & 75.6 & 58.7 & 26.6 & 19.7\\
    & RIENet & \checkmark & 46.2 & 83.3 & 73.2 & 43.5 & 19.8 & 11.1 \\
    & \cellcolor{mylightblue} Ours & \cellcolor{mylightblue} \checkmark & \cellcolor{mylightblue} \textbf{62.6} & \cellcolor{mylightblue} \textbf{97.5} & \cellcolor{mylightblue} \textbf{84.6} & \cellcolor{mylightblue} \textbf{62.6} & \cellcolor{mylightblue} \textbf{37.8} & \cellcolor{mylightblue} \textbf{30.2} \\
    \bottomrule
  \end{tabular}
\end{table}
\subsection{Performance Comparison with State-of-the-Art}
Quantitative results are presented in Table~\ref{tbl:overallperf}. Our method outperforms existing unsupervised approaches and achieving state-of-the-art performance across all datasets and demonstrates superior generalizability. Notably, our unsupervised approach maintains competitive performance compared to supervised methods and even surpasses them in distant scenarios, highlighting its potential for real-world application.
\paragraph{Overall Performance}
Compared to existing methods, our approach excels in performance. RIENet, an end-to-end unsupervised registration method for outdoor scenes, exhibits suboptimal performance, particularly in low-overlap scenarios and environments with low LiDAR resolution, such as \nuscnenes{}. Both EYOC and \mymethod{} adopt a teacher-student framework for unsupervised training. However, our method demonstrates superior accuracy across all evaluation metrics, overcoming challenges associated with pseudo-label discovery and the absence of feature-space knowledge transfer in EYOC.
\paragraph{Generalizability}
We assess generalizability on \nuscnenes{} using weights trained on \kitti{}. Variations in LiDAR resolutions between \nuscnenes{} and \kitti{} may lead to different point densities, potentially degrading extracted features. Compared to existing unsupervised methods, our approach exhibits superior generalizability to unseen datasets. This superiority can be attributed to \mymethod's design, which learns density-invariant features.

\subsection{Analysis}
\label{subsec:analysis}

\begin{wraptable}{r}{0.35\textwidth}
  \centering
  \renewcommand{\arraystretch}{0.8}
\caption{Different Pose Estimators in \fgcm}
  \label{tbl:poseest}
  \begin{tabular}{ccc}
    \toprule
    Pose  & tIR@1\textsuperscript{st} & Time \\
    Estimators & Epoch & (s) \\
    \midrule
    PointDSC & \textbf{81.3} & 1.13 \\
    MAC & 80.1 & 28.2\\
    FastMAC & 79.3 & \textbf{0.67}\\
    \midrule
    \scpcr & \underline{81.2} & \underline{0.75} \\
    \bottomrule
  \end{tabular}  
\end{wraptable}
\paragraph{Different Choices of Pose Estimator in \fgcm}
We contend that the robustness and efficacy of \fgcm{} are not contingent upon a specific pose estimator. 
To substantiate this claim, we conduct experiments employing various robust pose estimators within \fgcm{}. 
The results are detailed in Table~\ref{tbl:poseest}.
For different robust pose estimators in \fgcm{} module, we experiment PointDSC\citep{bai2021pointdsc}\footnote{We directly use their official weights for evaluation.}, MAC\cite{zhang20233d}, FastMAC\cite{zhang2024fastmac} and \scpcr\citep{chen2022sc2}.
The results demonstrate that the effectiveness of \fgcm{} is agnostic to choices of pose estimators, despite marginal performance discrepancies are observed.
Given the iterative nature of \fgcm{}, the efficiency of pose estimators holds paramount importance, 
as the module's runtime is proportional to pose estimation time. 
We choose \scpcr\citep{chen2022sc2} for \fgcm{} by default due to its superior balance in performance and efficiency.
\begin{figure}[htb]
  \centering
  \includegraphics[width=\textwidth]{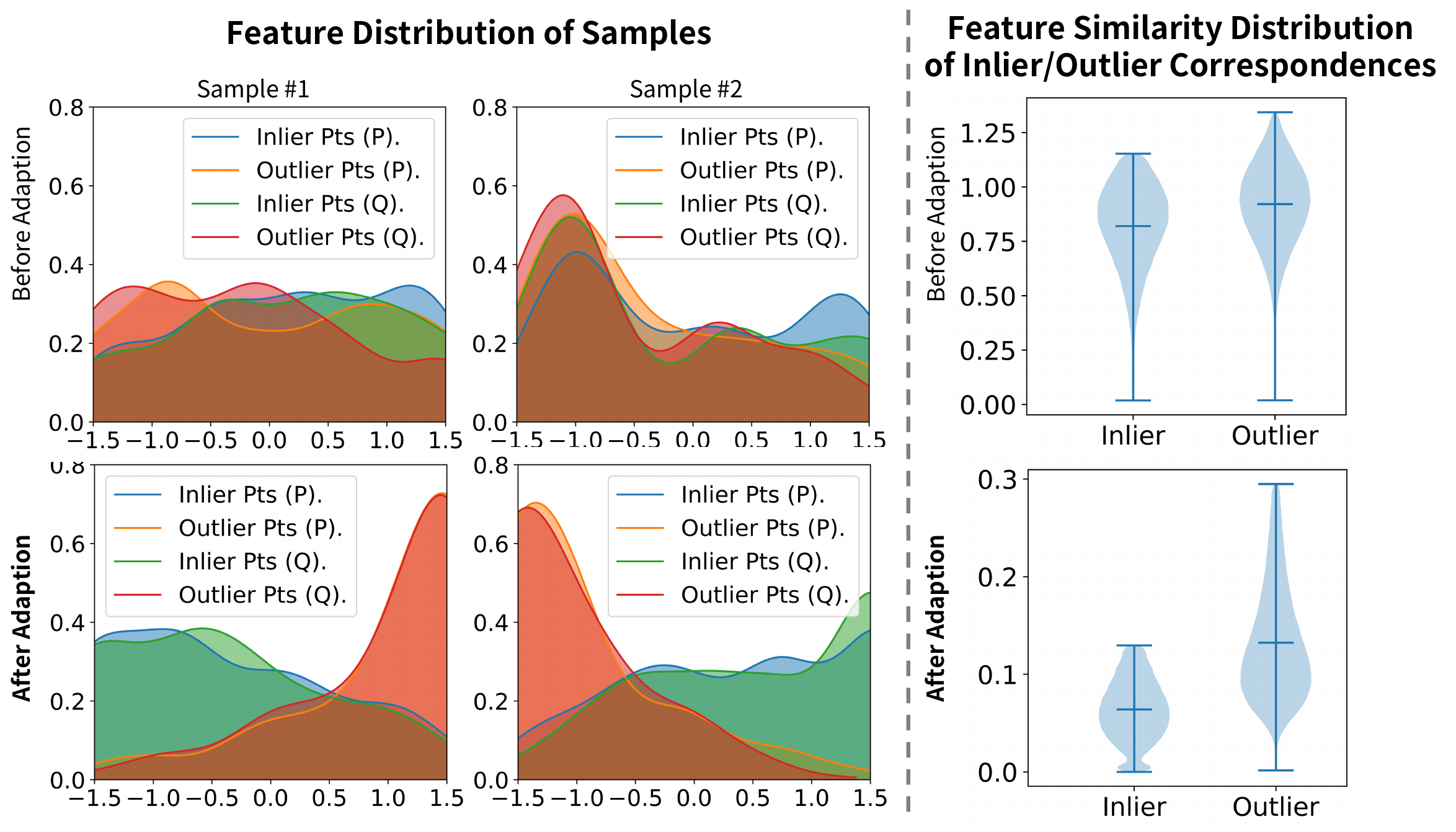}
  \caption{\textbf{Before v.s. After Self-Adaption in \fgcm}:
  Point-wise Feature \& Correspondence-wise Similarity Distribution indicate that the self-adaption results in more discriminative features. }
  \label{fig:adaption}
\end{figure}
\paragraph{Effectiveness of Self-Adaption for Discriminative Features.}
To further understand the effectiveness of self-adaption in \fgcm, we visualize the \emph{point-level feature distribution} and \emph{correspondence-level similarity distribution} in Fig.~\ref{fig:adaption} (Please refer to the Appendix for implementation details.).
The two representative samples are taken from \kitti{} dataset. 
In Fig.~\ref{fig:adaption}, the smaller overlap regions of point-level feature distribution between points from inliers and outliers indicate the features of inliers and outliers distribute more distant, and thus, the features are more discriminative.
For correspondence-level similarity, inlier similarity should be distinct from outlier similarity to effectively differentiate between the two.
With the self-adaption in \fgcm, the data-specific teacher produces more discriminative features, resulting in a less noisy feature space conducive to the subsequent feature-based approach employed in \fgcm.

\subsection{Ablation Study}
We conduct ablation studies to evaluate the efficacy of \mymethod{} on the \kitti{} dataset. We present mRR and registration errors in a distant scenario where \(d\!\!\in\!\![40,50)\). Various alternative configurations of \mymethod{} are compared in Table~\ref{tbl:ablation}. 
Our method exhibits superior performance compared to alternatives, demonstrating the effectiveness of our design. 
This superiority may be attributed to the features-geometry coherence: 
With \fgcm{}, correspondences in pseudo-labels possess discriminative features, facilitating effective knowledge transfer in feature space using \atl{}. 
\begin{wraptable}{r}{0.55\textwidth}
  \renewcommand{\arraystretch}{0.8}
  \centering
  \caption{\textbf{Ablation Study of \mymethod.} S.T.I denotes \textbf{s}ynthetic \textbf{t}eacher \textbf{i}nitialization. 
  PBSA and FGC denote \pbsafull{} and \fgcfull, respectively}
  \label{tbl:ablation}
  \begin{tabular}{cp{1cm}cccc}
    \toprule
    \multirow{2}{*}{Methods} & {tIR@1\textsuperscript{st}} & \multirow{2}{*}{mRR} & \multicolumn{3}{c}{\(d\in\left[40,50\right)\)} \\ 
    \cmidrule{4-6}
    & Epoch & & RR & RRE & RTE \\
    \midrule
    Full & \textbf{81.2} & \textbf{84.0} & \textbf{54.2} & \textbf{1.1} & \textbf{0.54}  \\
    \midrule
    w/o \atl{} & 80.3 & 83.5 & 53.7 & 1.3 & 0.58\\
    w/o PBSA & 43.3 & 80.9 & 50.2 & 1.7 & 0.79 \\
    w/o FGC & 67.6 & 82.8 & 52.7 & 1.4 & 0.61\\
    \midrule
    w/o \mds{} & 81.2 & 82.7 & 52.3 & 1.3 & 0.71\\
    w/o S.T.I &  71.9 & 83.7  & 53.7 & 1.2 & 0.55\\
    \bottomrule
  \end{tabular}
  \vspace{-0.3cm}
\end{wraptable}
Additionally, \mds{} significantly enhances performance in distant scenarios. 
The combination of \pbsafull{} and \fgcfull{} in the \fgcm{} module yields more substantial improvement than using either alone. The removal of \pbsafull{} marginally degrades the quality of pseudo-labels, emphasizing the importance of denoising the feature space. When synthetic teacher initialization is removed (w/o S.T.I), we employed the same way as EYOC to pretrain the teacher. We find that synthetic teacher initialization greatly enhances the initial teacher's performance. Please refer to the Appendix for more qualitative results on generated synthetic pairs.

\section{Conclusion}
\label{sec:concl}
In this paper, we present \mymethod{}, a novel unsupervised method for point cloud registration that integrates low-level geometric and high-level contextual information for reliable pseudo-labels. Our method introduces \fgcmfull{} for dynamic teacher self-adaption and robust pseudo-label mining based on both feature and geometric spaces. Then, we propose \mdsfull{} to learn density-invariant features. We also introduce \atlfull{} for effective contrastive learning using anchors. Extensive experiments on two large-scale outdoor datasets validate our method's efficacy. Despite being unsupervised, it achieves results comparable to state-of-the-art supervised methods and surpasses existing unsupervised methods, particularly in distant scenarios. Furthermore, our approach exhibits superior generalizability to unseen datasets.
\paragraph{Limitations.}
The main limitations of the proposed method are twofold:
\begin{itemize}[leftmargin=10pt]
    \item Our method is subject to the quality of the teacher. 
If the teacher is inaccurate, the feature space may become too noisy, potentially impeding \fgcfull{} in \fgcm, especially in distant scenarios. One potential remedy is to devise a more robust strategy for initializing the teacher.
    \item Our method is slightly slower to obtain pseudo-labels compared to existing efforts\citep{liu2024extend} due to the 
proposed iterative method used in \fgcm{} of the \fgcm{} module. Future work may involve devising a more efficient strategy for mining pseudo-labels.
\end{itemize}
\section{Acknowledgements}
This work was supported in part by the National Key R\&D Program of China under Grant 2021YFF0704600, the Fundamental Research Funds for the Central Universities (No. 20720220064).

\bibliography{unsupreg_nips}

\end{document}